\documentclass[graybox]{svmult}

\usepackage{mathptmx}       
\usepackage{helvet}         
\usepackage{courier}        
\usepackage{makeidx}         
\usepackage{graphicx}        
\usepackage{multicol}        
\usepackage[bottom]{footmisc}

\makeindex             

\usepackage{amsfonts,amsmath,url}

\begin{document}

\title*{Constrained variable clustering and the best basis problem in
  functional data analysis}
\author{Fabrice Rossi and Yves Lechevallier}
\institute{F. Rossi 
\at Institut T\'el\'ecom,  T\'el\'ecom ParisTech, LTCI - UMR CNRS 5141 
46, rue Barrault, 75013 Paris, France,
\email{Fabrice.Rossi@telecom-paristech.fr}
\and
Y. Lechevallier
\at Projet AxIS, INRIA, Domaine de Voluceau, Rocquencourt, B.P. 105,  78153 Le
Chesnay Cedex, France, \email{Yves.Lechevallier@inria.fr}}
%
%
\maketitle

\abstract{Functional data analysis involves data described by regular
  functions rather than by a finite number of real valued variables. While
  some robust data analysis methods can be applied directly to the very high
  dimensional vectors obtained from a fine grid sampling of functional data,
  all methods benefit from a prior simplification of the functions that
  reduces the redundancy induced by the regularity. In this paper we propose
  to use a clustering approach that targets variables rather than individual
  to design a piecewise constant representation of a set of functions. The
  contiguity constraint induced by the functional nature of the variables
  allows a polynomial complexity algorithm to give the optimal solution.}

\section{Introduction}
Functional data \cite{RamsaySilverman97} appear in applications in which
objects to analyse display some form of variability. In spectrometry, for
instance, samples are described by spectra: each spectrum is a mapping from
wavelengths to e.g., transmittance\footnote{In spectrometry, transmittance is
  the fraction of incident light at a specified wavelength that passes through
  a sample.}. Time varying objects offer a more general example: when the
characteristics of objects evolve through time, a loss free representation
consists in describing these characteristics as functions that map time to
real values.

In practice, functional data are given as high dimensional vectors (e.g., more
than 100 variables) obtained by sampling the functions on a fine grid. For
smooth functions (for instance in near infrared spectroscopy), this scheme
leads to highly correlated variables. While many data analysis methods can be
made robust to this type of problem (see, e.g.,
\cite{HastieBujaTibshirani1995} for discriminant analysis), all methods
benefit from a compression of the data \cite{OlssonEtAl1996Bsplines} in which
relevant and yet easy to interpret features are extracted from the raw
functional data.

There are well-known standard ways of extracting optimal features according to
a given criterion. For instance in unsupervised problems, the first $k$
principal components of a dataset give the best linear approximation of the
original data in $\mathbb{R}^k$ for the quadratic norm (see
\cite{RamsaySilverman97} for functional principal component analysis
(PCA)). In regression problems, the partial least-squares approach extracts
features with maximal correlation with a target variable (see also Sliced Inversion Regression methods \cite{FerreYao2003SIR}). The main drawback of
those approaches is that they extract features that are not easy to interpret:
while the link between the original features and the new ones is linear, it
is seldom sparse; an extracted feature generally depends on many original
features. 

A different line of thoughts is followed in the present paper: the
goal is to extract features that are easy to interpret in terms of the
original variables. This is done by approximating the original functions by
piecewise constant functions. We first recall in Section \ref{bestbasis} the
best basis problem in the context of functional data approximation. Section
\ref{constrainedclustering} shows how the problem can be recast in term of a
constrained clustering problem for which efficient solutions are
available. 

\section{Best basis for functional data}\label{bestbasis}
Let us consider $n$ functional data, $(s_i)_{1\leq i\leq n}$. Each $s_i$ is a
function from $[a,b]$ to $\mathbb{R}$, where $[a,b]$ is a fixed interval
common to all functions (more precisely, $s_i$ belongs to $L^2([a,b])$, the
set of square integrable functions on $[a,b]$). In terms of functional data,
linear feature extraction consists in choosing for each feature a linear
operator from $L^2([a,b])$ to $\mathbb{R}$. Equivalently, one can choose a
function $\phi$ from $L^2([a,b])$ and compute $\langle
s_i,\phi\rangle_{L^2}=\int_a^b\phi(x)s_i(x)dx$. In an unsupervised context, using
e.g., a quadratic error measure, choosing the $k$ best features consists in
finding $k$ orthonormal functions $(\phi_i)_{1\leq i\leq k}$ that minimise the
following quantity:
\begin{equation}
  \label{eq:QuadraticError}
\sum_{i=1}^n\left\|s_i-\sum_{j=1}^k\langle
  s_i,\phi_k\rangle_{L^2}\phi_k\right\|_{L^2}^2.
\end{equation}
The $(\phi_i)_{1\leq i\leq k}$ form an orthonormal basis of the subspace that
they span: the optimal set of such functions is therefore called the
\emph{best basis} for the original set of functions $(s_i)_{1\leq i\leq n}$. 

If the $\phi_k$ are unconstrained, the best basis is given by functional
PCA \cite{RamsaySilverman97}. However, in order for the corresponding
feature to be easy to interpret, the $\phi_k$ should have compact supports, the
simple case of $\phi_k=\mathbb{I}_{[u_k,v_k]}$ being the easiest to analyse
($\mathbb{I}_{[u,v]}(x)=1$ when $x\in[u,v]$ and 0 elsewhere). 

The problem of choosing an optimal basis among a set of bases has been studied
for some time in the wavelet community
\cite{CoifmanWickerhauser1992BestBasis,SaitoCoifman1995LocalBases}. In
unsupervised context, the best basis is obtained by minimizing the entropy of
the features (i.e., of the coordinates of the functions on the basis) in order
to enable compression by discarding the less important features. Following
\cite{OlssonEtAl1996Bsplines}, \cite{RossiEtAl06CilsBspline} proposes a
different approach, based on B-splines: a leave-one-out version of Equation
\eqref{eq:QuadraticError} is used to select the best B-splines basis. While
the orthonormal basis induced by the B-splines does not correspond to
compactly supported functions, the dependency between a new feature and the
original ones is still localized enough to allow easy
interpretation. Nevertheless both approaches have some drawbacks. Wavelet
based methods lead to compactly supported basis functions but the basis has to
be chosen in a tree structured set of bases. As a consequence, the support of a
basis function cannot be any sub-interval of $[a,b]$. The B-spline approach
suffers from a similar problem: the approximate supports have all the same
lengths leading either to a poor representation of some local details or to a
large number of basis functions.  

\section{Best basis via constrained clustering}\label{constrainedclustering}
\subsection{From best basis to constrained clustering}
The goal of the present paper is to select an optimal basis using only basis
functions of the form $\mathbb{I}_{(u,v)}$, without restriction on the possible
intervals among sub-interval of $[a,b]$\footnote{The notations $(u,v)$ is used
to include all the possible cases of open and close boundaries for the
considered intervals.}. Let us consider
$(\phi_j=\frac{1}{v_j-u_j}\mathbb{I}_{(u_j,v_j)})_{1\leq j\leq k}$ such an
orthonormal basis. We assume that the $((u_j,v_j))_{1\leq j\leq
  k}$ form a partition of $[a,b]$. Obviously, we have $\langle
\phi_j,s_i\rangle=\frac{1}{v_j-u_j}\int_{u_j}^{v_j}s_i(x)dx$, i.e., the
feature corresponding to $\phi_j$ is the mean value of $s_i$ on
$[u_j,v_j]$. In other words, $\sum_{j=1}^k\langle
s_i,\phi_k\rangle_{L^2}\phi_k$ is a piecewise constant approximation of $s_i$
(which is optimal according to the $L^2$ norm).

In practice, functional data are sampled on a fine grid with support points
$a\leq t_1<\ldots<t_m\leq b$, i.e., rather than observing the functions
$(s_i)_{1\leq i\leq n}$, one gets the vectors $(s_i(t_l))_{1\leq i\leq n,1\leq
  l\leq m}$ from $\mathbb{R}^m$. Then $\langle \phi_j,s_i\rangle$ can be
approximated by $\frac{1}{|I_j|}\sum_{l\in I_j}s_i(t_l)$ where $I_j$ is the
subset of indexes $\{1,...,m\}$ such that $t_l\in(u_j,v_j)\Leftrightarrow l\in
I_j$. Any partition of $((u_j,v_j))_{1\leq j\leq k}$ of $[a,b]$ corresponds to
a partition of $\{1,...,m\}$ in $k$ subsets $(I_j)_{1\leq j\leq k}$ that
satisfies an ordering constraint: if $r$ and $s$ belong to $I_j$ then any
integer $t\in[r,s]$ belongs also to $I_j$. Finding the best basis means for
instance minimizing the sum of squared errors given by Equation
\eqref{eq:QuadraticError} which can be approximated as follows
\begin{equation}
  \label{eq:QECluster}
\sum_{i=1}^n\sum_{j=1}^k\sum_{l\in I_j}\left(s_i(t_l)-\frac{1}{|I_j|}\sum_{u\in
  I_j}s_i(t_u)\right)^2=\sum_{j=1}^kQ(I_j),
\end{equation}
where
\begin{equation}
  \label{eq:Qinterval}
Q(I)=  \sum_{i=1}^n\sum_{l\in I}\left(s_i(t_l)-\frac{1}{|I|}\sum_{u\in I}s_i(t_u)\right)^2
\end{equation}
The second version of the error shows that it corresponds to an additive
quality measure of the partition of $\{1,...,m\}$ induced by the $(I_j)_{1\leq
  j\leq  k}$. Therefore, finding the best basis for the sampled functions is
equivalent to finding an optimal partition of $\{1,...,m\}$ with some ordering
constraints and according to an additive cost function. A suboptimal solution
to this problem, based on an ascending (agglomerative) hierarchical
clustering, is proposed in \cite{KrierEtAl2007CILSFDProj}. 

\subsection{Dynamic programming}
However, an optimal solution can be reached in a reasonable amount of time,
as pointed out in \cite{LechevallierContrainte1976}: when the quality
criterion of a partition is additive and when a total ordering constraint is
enforced, a dynamic programming approach leads to the optimal solution (this
is a generalization of the algorithm proposed by Bellman for a single function
in \cite{Stone1961,Bellman1961Function}; see also
\cite{AugerLawrence1989,JacksonEtAl2005} for 
rediscoveries/extensions of this early work). The
algorithm is simple and proceeds iteratively by computing $F(j,k)$ as the
value of the quality measure (from Equation \eqref{eq:QECluster}) of the best
partition in $k$ classes of $\{j,...,m\}$:
\begin{enumerate}
\item initialization: set $F(j,1)$ to $Q(\{j,\ldots,m\})$ for all $j$
\item iterate from $p=2$ to $k$:
  \begin{enumerate}
  \item for all $1\leq j\leq m-p+1$ compute 
\[
F(j,p)=\min_{j\leq l\leq m-p+1}Q(\{j,\ldots,l\})+F(l+1,p-1)
\]
  \end{enumerate}
\end{enumerate}
The minimizing index $l=l(j,p)$ is kept for all $j$ and $p$. This allows to
reconstruct the best partition by backtracking from $F(1,k)$: the first class
of the partition is $\{1,\ldots,l(1,k)\}$, the second
$\{l(1,k)+1,\ldots,l(l(1,k)+1,k-1)\}$, etc.  A similar algorithm was used to
find an optimal approximation of a single function in
\cite{Bellman1961Function,LechevallierContrainte1990}. Another related work is
\cite{HugueneyEtAlSFDS2006} which provides simultaneously a functional
clustering and a piecewise constant approximation of the prototype functions. 

The internal loop runs $O(km^2)$ times. It uses the values $Q(\{j,\ldots,l\})$
for all $j\leq l$. Those quantities can be computed prior to the search for
the optimal partition, using for instance a recursive variance computation
formula, leading to a cost in $O(nm^2)$. More precisely, we are interested in
\begin{equation}  \label{eq:Qjl}
Q_{i,j,l}=\sum_{r=j}^l(s_i(t_r)-M_{i,j,l})^2,
\end{equation}
where
\begin{equation}  \label{eq:Mjl}
M_{i,j,l}=\frac{1}{l-j+1}\sum_{u=j}^ls_i(t_u).
\end{equation}
For a fixed function $s_i$, the $M_{i,j,l}$ and $Q_{i,j,l}$ are computed and
stored in two $m\times m$ arrays, according to the following algorithm:
\begin{enumerate}
\item initialisation: set $M_{i,j,j}=s_i(t_j)$ and $Q_{i,j,j}=0$ for all
  $j\in\{1,\ldots,m\}$
\item compute $M_{i,1,j}$ and $Q_{i,1,j}$ for $j>1$ recursively with:
  \begin{eqnarray*}
M_{i,1,j}&=&\frac{1}{j}\left((j-1)M_{i,1,j-1}+s_i(t_j)\right)\\
Q_{i,1,j}&=&Q_{i,1,j-1}+\frac{j}{j-1}(s_i(t_j)-M_{i,1,j})^2\\
  \end{eqnarray*}
\item compute $M_{i,j,l}$ and $Q_{i,j,l}$ for $l>j>1$ recursively with:
  \begin{eqnarray*}
M_{i,j,l}&=&\frac{1}{l-j+1}\left((l-j+2)M_{i,j-1,l}-s_i(t_{j-1})\right)\\
Q_{i,j,l}&=&Q_{i,j-1,l}-\frac{l-j+1}{l-j+2}(s_i(t_{j-1})-M_{i,j,l})^2\\
  \end{eqnarray*}
\end{enumerate}
This algorithm is applied to each function leading to a total cost of
$O(nm^2)$ with a $O(m^2)$ storage. The full algorithm has therefore a
complexity of $O((n+k)m^2)$. 

\subsection{Extensions}
As pointed out in \cite{LechevallierContrainte1976}, the previous scheme can
be used for any additive quality measure. It is therefore possible to use
e.g., a piecewise linear approximation of the functions on a sub-interval
rather than a constant approximation (this is the original problem studied in
\cite{Bellman1961Function} for a single function). However, additivity is a
stringent restriction. In the case of a piecewise linear approximation for
instance, it prevents the introduction of continuity conditions: if one
searches for the best continuous piecewise linear approximation of a function,
then the optimized criterion is no more additive (this is in fact the case for
all spline smoothing approaches expect the piecewise constant ones).

In addition, for the general case of an arbitrary quality measure $Q$ there
might be no recursive formula for evaluating $Q$. In this case, the cost of
computing the needed quantities might exceed $O(nm^2)$ and reach $O(nm^3)$ or
more, depending on the exact definition of $Q$.

That said, the particular case of leave-one-out is quite interesting. Indeed
when the studied functions are noisy, it is important to rely on a good
estimate of the approximation error to avoid overfitting the best basis to the
noise. It is straightforward to show that the leave-one-out (l.o.o.) estimate
of the total error from equation \eqref{eq:QECluster} is given by
\begin{equation}
  \label{eq:QECluster:loo}
\sum_{i=1}^n\sum_{j=1}^k\sum_{l\in I_j}\left(\frac{|I_j|}{|I_j|-1}\right)^2\left(s_i(t_l)-\frac{1}{|I_j|}\sum_{u\in I_j}s_i(t_u)\right)^2,  
\end{equation}
when l.o.o. is done on the sampling points of the functions. This is an additive
quality measure which can be computed using from the $Q_{i,j,l}$, that is in
an efficient recursive way. As shown above, the piecewise constant
approximation with $k$ segments is obtained via the computation of the best
approximation for all $l$ in $\{1,\ldots,k\}$. It is then possible to choose
the best $l$ based on the leave-one-out error estimate at the same cost as the
one needed to compute the best approximation for the maximal value of
$l$. This leads to two variants of the algorithm. In the first one, the
standard algorithm is applied to compute all the best bases and the best
number of segments is chosen via the l.o.o. error estimate (which can be readily
computed once the best basis is known). In the second one, we compute the best
basis directly according to the l.o.o. error estimate, leveraging its additive
structure. It is expected that this second solution will perform better in
practice, as it constrains the best basis to be reasonable (see Section
\ref{experiments} for an experimental validation). For instance, it
will never select an interval with only one point whereas this could be the
case for the standard solution. As a consequence, the standard solution will
likely produce bases with rather bad leave-one-out performances and tend to
select a too small number of segments (see Section \ref{experiments} for an
example of this behavior). 

\begin{figure}[htbp]
  \centering
\includegraphics[angle=270,width=0.9\textwidth]{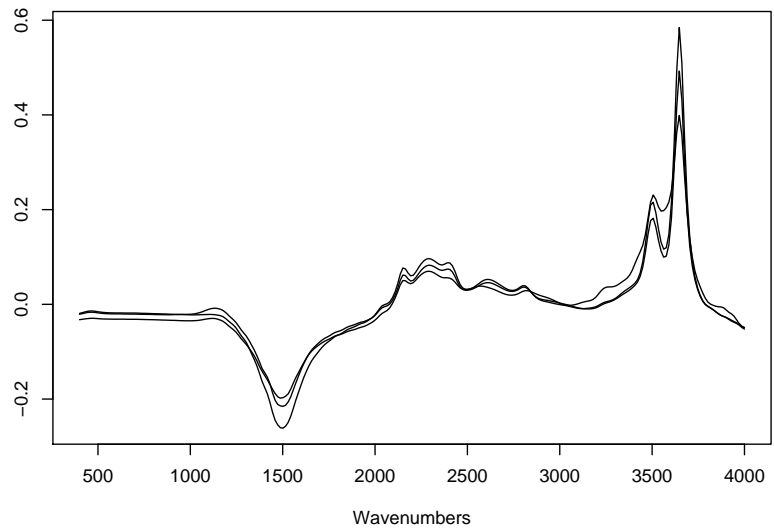}  
  \caption{Three spectra from the Wine dataset}
  \label{fig:wine}
\end{figure}

\section{Experiments}\label{experiments}
\begin{figure}[htbp]
  \centering
\includegraphics[angle=270,width=0.90\textwidth]{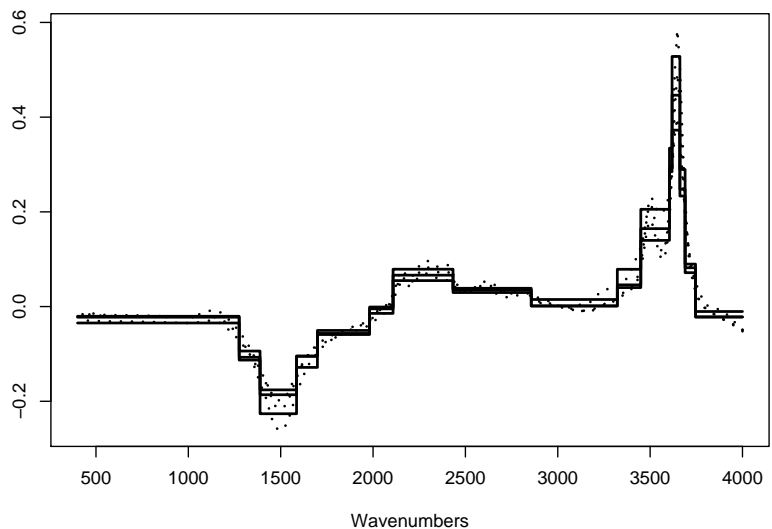} 
  \caption{Example of the optimal approximation results for  16 clusters on
    the Wine dataset}
  \label{fig:wine:approx:good:opti}
\end{figure}

\begin{figure}[htbp]
  \centering
\includegraphics[angle=270,width=0.90\textwidth]{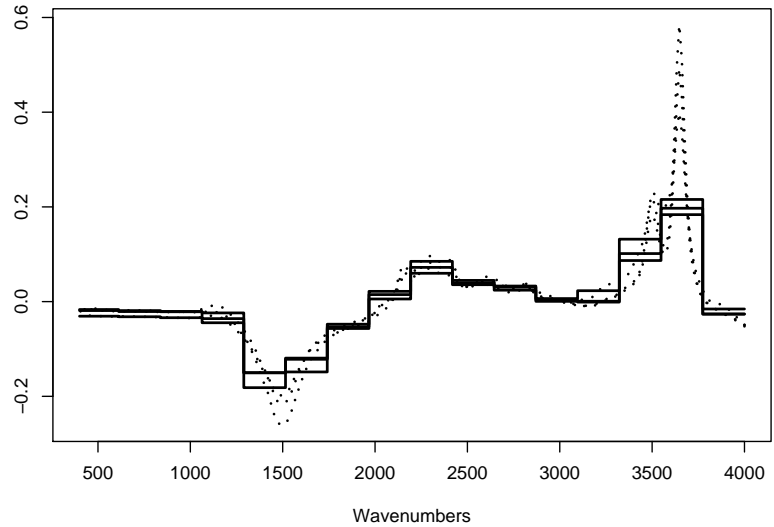}

  \caption{Example of the uniform approximation results for  16 clusters on
    the Wine dataset}
  \label{fig:wine:approx:good:uniform}
\end{figure}

We illustrate the algorithm on the Wine dataset\footnote{This dataset is
  provided by Prof. Marc Meurens, Universit\'e catholique de Louvain, BNUT
  unit, and available at
  \url{http://www.ucl.ac.be/mlg/index.php?page=DataBases}.} which consists in
124 spectra of wine samples recorded in the mid infrared range at 256
different wavenumbers\footnote{The wavenumber is the inverse of the
  wavelength.} between 4000 and 400 cm$^{-1}$. Spectra number 34, 35 and 84 of
the learning set of the original dataset have been removed as they are
outliers. As shown on Figure \ref{fig:wine} the function approximation problem
is interesting as the smoothness of the spectrum varies along the spectral
range and an optimal basis will obviously not consist in functions with
supports of equal size. Figure \ref{fig:wine:approx:good:opti} shows an
example of the best basis obtained by the proposed approach for $k=16$
clusters, while Figure \ref{fig:wine:approx:good:uniform} gives the suboptimal
solution obtained by a basis with equal length intervals (as used in
\cite{RossiEtAl06CilsBspline}). The uniform length approach is clearly unable
to pick up details such as the peak on the right of the spectra. The total
approximation error (equation \eqref{eq:QECluster}) is reduced from $62.66$
with the uniform approach to $7.74$ with the optimal solution. On the same
dataset, the greedy ascending hierarchical clustering approach proposed in
\cite{KrierEtAl2007CILSFDProj} reaches a total error of $8.55$ for a similar
running time of the optimal approach proposed in the present paper.

To test the leave-one-out approach, we have first added a Gaussian noise with
$0.04$ standard deviation (the functions take values in
$[-0.265,0.581]$). Then we look for the best basis up to 64 segments. As
expected, the total approximation error decreases with the number of segments
and would therefore lead to a best basis with 64 segments. Moreover, as
explained in the previous Section, the bases are not controlled by a
l.o.o. error estimate. As a consequence, the optimization leads very quickly
to basis with very small segments (starting at $k=12$, there is at least one
segment with only one sample point in it). Therefore, the l.o.o. error
estimate applied to this set of bases selects a quite low number of segments,
namely $k=11$.  When the bases are optimized according to the l.o.o. error
estimate, the behavior is more smooth in the sense that small segments are
always avoided. The minimum value of the l.o.o. estimate leads to the
selection of $k=20$ segments.

\begin{table}[htbp]
  \centering
  \begin{tabular}{lcc}\hline\noalign{\smallskip}
Basis & Noisy data & Real spectra\\\noalign{\smallskip}\svhline\noalign{\smallskip}
$k=64$ (standard approach) & 37.28 & 14.35 \\
$k=11$ (l.o.o. after the standard approach) & 63.19 & 17.35 \\
$k=20$ (full l.o.o.) & 54.07 & 12.07 \\\noalign{\smallskip}\hline
  \end{tabular}
  \caption{Total squared errors for the Wine dataset with noise }
  \label{table:wine:noise}
\end{table}

\begin{figure}[htbp]
  \centering
\includegraphics[angle=270,width=0.90\textwidth]{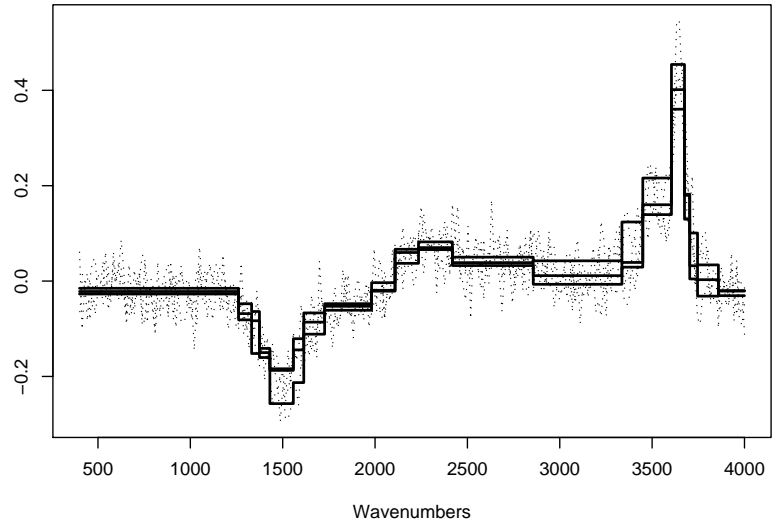} 

  \caption{Best basis selected by leave-one-out with the standard approach
    combined with loo}
  \label{fig:wine:noise:approx:loo}
\end{figure}

\begin{figure}[htbp]
  \centering
\includegraphics[angle=270,width=0.90\textwidth]{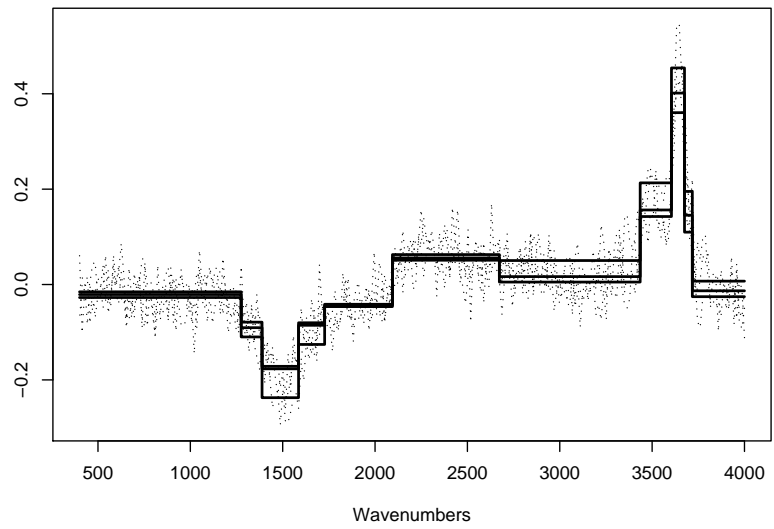}

  \caption{Best basis selected by leave-one-out with the full loo approach}
  \label{fig:wine:noise:approx:full}
\end{figure}

Table \ref{table:wine:noise} summarizes the results by displaying the total
approximation error on the noisy spectra and the total approximation error on
the original spectra (the ground truth) for the three alternatives. The full
l.o.o. approach leads clearly to the best results, as illustrated on Figures
\ref{fig:wine:noise:approx:loo} and \ref{fig:wine:noise:approx:full}.

Those experiments show that the proposed approach is flexible and provides an
efficient way to get an optimal basis for a set of functional data. We are
currently investigating supervised extensions of the approach following
principles from \cite{FrancoisEtAl2008Agrostat}.

\section*{Acknowledgement}
The authors thank the anonymous reviewer for the detailed and constructive comments
that have significantly improved this paper.


\begin{thebibliography}{10}
\providecommand{\url}[1]{{#1}}
\providecommand{\urlprefix}{URL }
\expandafter\ifx\csname urlstyle\endcsname\relax
  \providecommand{\doi}[1]{DOI~\discretionary{}{}{}#1}\else
  \providecommand{\doi}{DOI~\discretionary{}{}{}\begingroup
  \urlstyle{rm}\Url}\fi

\bibitem{AugerLawrence1989}
Auger, I.E., Lawrence, C.E.: Algorithms for the optimal identification of
  segment neighborhoods.
\newblock Bulletin of Mathematical Biology \textbf{51}(1), 39--54 (1989)

\bibitem{Bellman1961Function}
Bellman, R.: On the approximation of curves by line segments using dynamic
  programming.
\newblock Communication of the ACM \textbf{4}(6), 284 (1961).
\newblock \doi{http://doi.acm.org/10.1145/366573.366611}

\bibitem{CoifmanWickerhauser1992BestBasis}
Coifman, R.R., Wickerhauser, M.V.: Entropy-based algorithms for best basis
  selection.
\newblock IEEE Transactions on Information Theory \textbf{38}(2), 713--718
  (1992)

\bibitem{FerreYao2003SIR}
Ferr{\'e}, L., Yao, A.F.: Functional sliced inverse regression analysis.
\newblock Statistics \textbf{37}(6), 475--488 (2003)

\bibitem{FrancoisEtAl2008Agrostat}
Fran{\c c}ois, D., Krier, C., Rossi, F., Verleysen, M.: Estimation de
  redondance pour le clustering de variables spectrales.
\newblock In: Actes des 10{\`e}mes journ{\'e}es Europ{\'e}ennes Agro-industrie
  et M{\'e}thodes statistiques (Agrostat 2008), pp. 55--61. Louvain-la-Neuve,
  Belgique (2008)

\bibitem{HastieBujaTibshirani1995}
Hastie, T., Buja, A., Tibshirani, R.: Penalized discriminant analysis.
\newblock Annals of Statistics \textbf{23}, 73--102 (1995)

\bibitem{HugueneyEtAlSFDS2006}
Hugueney, B., H{\'e}brail, G., Lechevallier, Y.: R{\'e}duction de s{\'e}ries
  temporelles par classification et segmentation.
\newblock In: Actes des 38i{\`e}mes Journ{\'e}es de Statistique de la SFDS.
  Clamart, France (2006)

\bibitem{JacksonEtAl2005}
Jackson, B., Scargle, J., Barnes, D., Arabhi, S., Alt, A., Gioumousis, P.,
  Gwin, E., Sangtrakulcharoen, P., Tan, L., Tsai, T.T.: An algorithm for
  optimal partitioning of data on an interval.
\newblock IEEE Signal Processing Letters \textbf{12}(2), 105--108 (2005)

\bibitem{KrierEtAl2007CILSFDProj}
Krier, C., Rossi, F., François, D., Verleysen, M.: A data-driven functional
  projection approach for the selection of feature ranges in spectra with ica
  or cluster analysis.
\newblock Chemometrics and Intelligent Laboratory Systems \textbf{91}(1),
  43--53 (2008)

\bibitem{LechevallierContrainte1976}
Lechevallier, Y.: Classification automatique optimale sous contrainte d'ordre
  total.
\newblock Rapport de recherche 200, IRIA (1976)

\bibitem{LechevallierContrainte1990}
Lechevallier, Y.: Recherche d'une partition optimale sous contrainte d'ordre
  total.
\newblock Rapport de recherche RR-1247, INRIA (1990).
\newblock \url{http://www.inria.fr/rrrt/rr-1247.html}

\bibitem{OlssonEtAl1996Bsplines}
Olsson, R.J.O., Karlsson, M., Moberg, L.: Compression of first-order spectral
  data using the b-spline zero compression method.
\newblock Journal of Chemometrics \textbf{10}(5--6), 399--410 (1996)

\bibitem{RamsaySilverman97}
Ramsay, J., Silverman, B.: Functional Data Analysis.
\newblock Springer Series in Statistics. Springer Verlag (1997)

\bibitem{RossiEtAl06CilsBspline}
Rossi, F., Fran{\c c}ois, D., Wertz, V., Verleysen, M.: Fast selection of
  spectral variables with b-spline compression.
\newblock Chemometrics and Intelligent Laboratory Systems \textbf{86}(2),
  208--218 (2007)

\bibitem{SaitoCoifman1995LocalBases}
Saito, N., Coifman, R.R.: Local discriminant bases and their applications.
\newblock Journal of Mathematical Imaging and Vision \textbf{5}(4), 337--358
  (1995)

\bibitem{Stone1961}
Stone, H.: Approximation of curves by line segments.
\newblock Math. Comput. \textbf{15}, 40--47 (1961)

\end{thebibliography}

\end{document}